\documentclass[fleqn,10pt,twocolumn]{SICE18}
\usepackage{float}
\usepackage{url}
\usepackage{graphicx}

\makeatletter

\providecommand{\tabularnewline}{\\}

\makeatother
\title{Synthesizing Chemical Plant Operation Procedures \\
using Knowledge, Dynamic Simulation and Deep Reinforcement Learning}
\author{Shumpei Kubosawa${}^{1,2,\dagger}$, Takashi Onishi${}^{1,2}$ and Yoshimasa Tsuruoka${}^{1,3}$}
\speaker{Shumpei Kubosawa}

\affils{${}^{1}$NEC-AIST AI Cooperative Research Laboratory, AIST, Tokyo, Japan\\
(Tel: +81-50-3362-9306; E-mail: \{shumpei.kubosawa,takashi.onishi,yoshimasa.tsuruoka\}@aist.go.jp)\\
${}^{2}$Security Research Laboratories, NEC Corporation, Kanagawa, Japan\\
${}^{3}$Department of Information and Communication Engineering, The University of Tokyo, Tokyo, Japan\\
}
\abstract{%
Chemical plants are complex and dynamical systems consisting of many
components for manipulation and sensing, whose state transitions
depend on various factors such as time, disturbance, and operation
procedures.
For the purpose of supporting human operators of chemical plants, we
are developing an AI system that can semi-automatically synthesize
operation procedures for efficient and stable operation.
Our system can provide not only appropriate operation procedures but
also reasons why the procedures are considered to be valid. This is
achieved by integrating automated reasoning and deep reinforcement
learning technologies with a chemical plant simulator and external
knowledge.
Our preliminary experimental results demonstrate that it can
synthesize a procedure that achieves a much faster recovery from a
malfunction compared to standard PID control.
}

\keywords{%
Process Optimization, Dynamic Simulation, Automated Reasoning, Deep Reinforcement Learning
}


\begin{document}
\maketitle

\section{Introduction}

Chemical plants are complex and dynamical systems consisting of many
components for manipulation and sensing. Operation of chemical plants
is thus not straight-forward and requires skilled and experienced
operators. Such operators are, however, rapidly retiring in the aging
population of Japan, and the industry is likely to face a serious
shortage of experienced and skilled operators in the near future.

To tackle this problem, we develop an artificial intelligence (AI)
system that can assist human operators of chemical plants. More specifically,
our AI system is designed to suggest appropriate operation procedures
to human operators and present reasons why those procedures are considered
to be appropriate. The ability of explaining the reasoning process
of AI is especially important since human operators would not adopt
a suggested operation procedure unless they are convinced with its
validity. 

Our system is based on simulation models for chemical plants and the
combination of deep learning and reinforcement learning, which is
often called deep reinforcement learning and has recently been applied
to control problems in various domains such as robotics \cite{Lillicrap}.
Although deep reinforcement learning has proven successful in many
control problems, finding appropriate operation procedures for chemical
plants is a highly challenging problem due to the vast search space
of possible operation procedures that result from many continuous
control variables in plant simulation models. To narrow down the search
space, our system uses qualitative knowledge derived from manuals
and focus only on promising operation procedures.

Our approach differs from previous work on automatic synthesis of
operation procedures in some significant ways. While the models used
in previous studies \cite{AOPS,Gofuku} are functional or qualitative,
our system employs a quantitative and dynamic model for simulation
and thus can produce precise procedures with numerical control values.
Quantitative and dynamic simulation models have been used for the
purpose of training human operators \cite{Kano}, but not for the
purpose of finding appropriate operation procedures. Numerical approaches
based on model predictive control (MPC) could be used to find quantitative
operation procedures \cite{Kano}, but MPC requires the model to be
differentiable and is not as flexible as deep reinforcement learning
in terms of problem formulation.
\begin{figure}[t]
\begin{centering}
\includegraphics[clip,width=7.5cm]{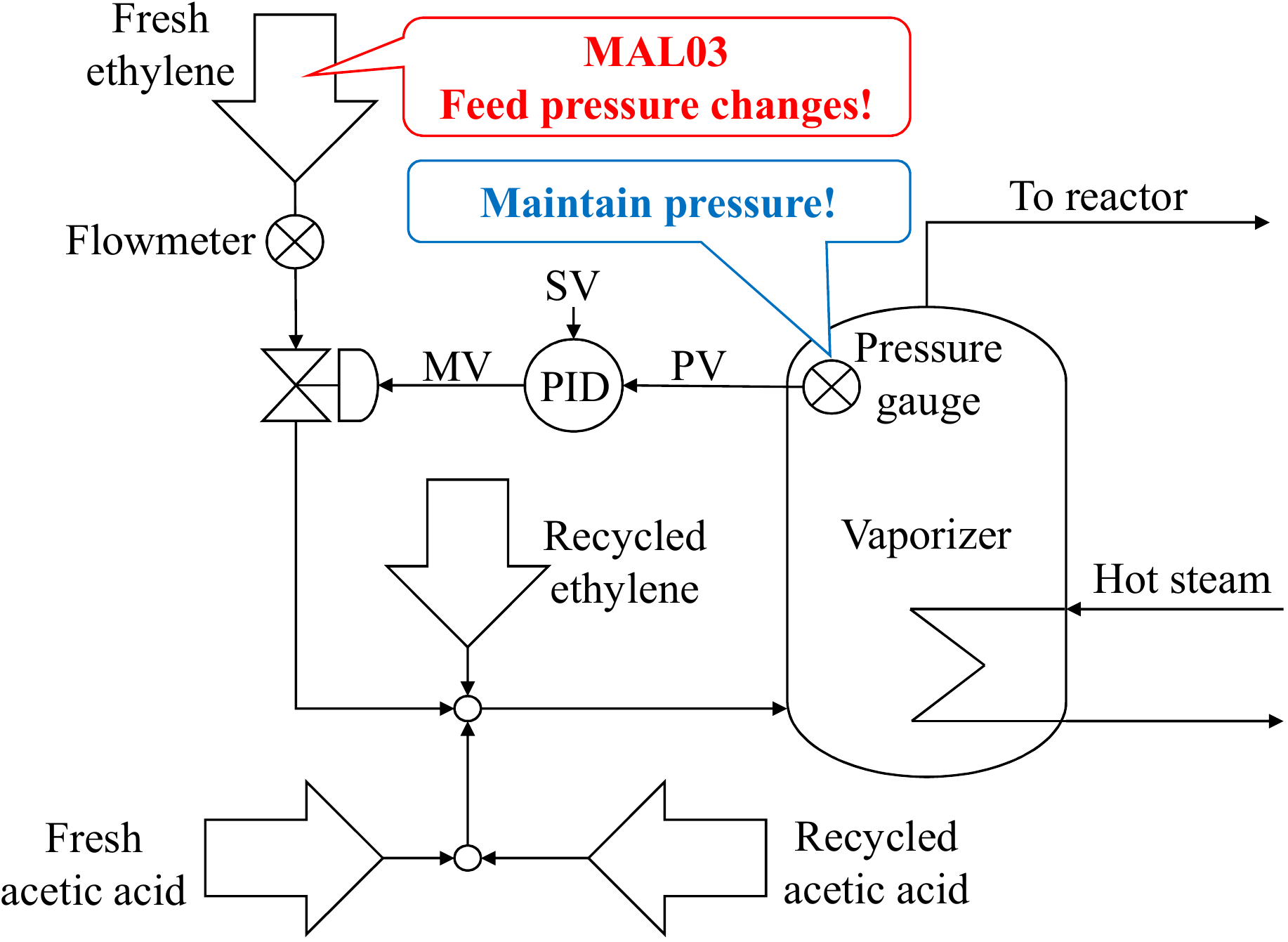}
\par\end{centering}
\caption{P\&ID of the raw material feed section of the VAM plant model focused
on MAL03 malfunction \label{fig:pnid}}
\end{figure}

In this paper, we use a simple model and a malfunction scenario shown
in Fig. \ref{fig:pnid} to evaluate the effectiveness of our approach.
Our system computes possible operation procedures for a raw material
feed section of a vynil acetate monomer (VAM) plant to recover from
a malfunction of feed pressure. Our preliminary experiments demonstrate
that it can synthesize a procedure that achieves a much faster recovery
from a malfunction compared to standard PID control. 
\begin{figure}[t]
\begin{centering}
\includegraphics[width=8cm]{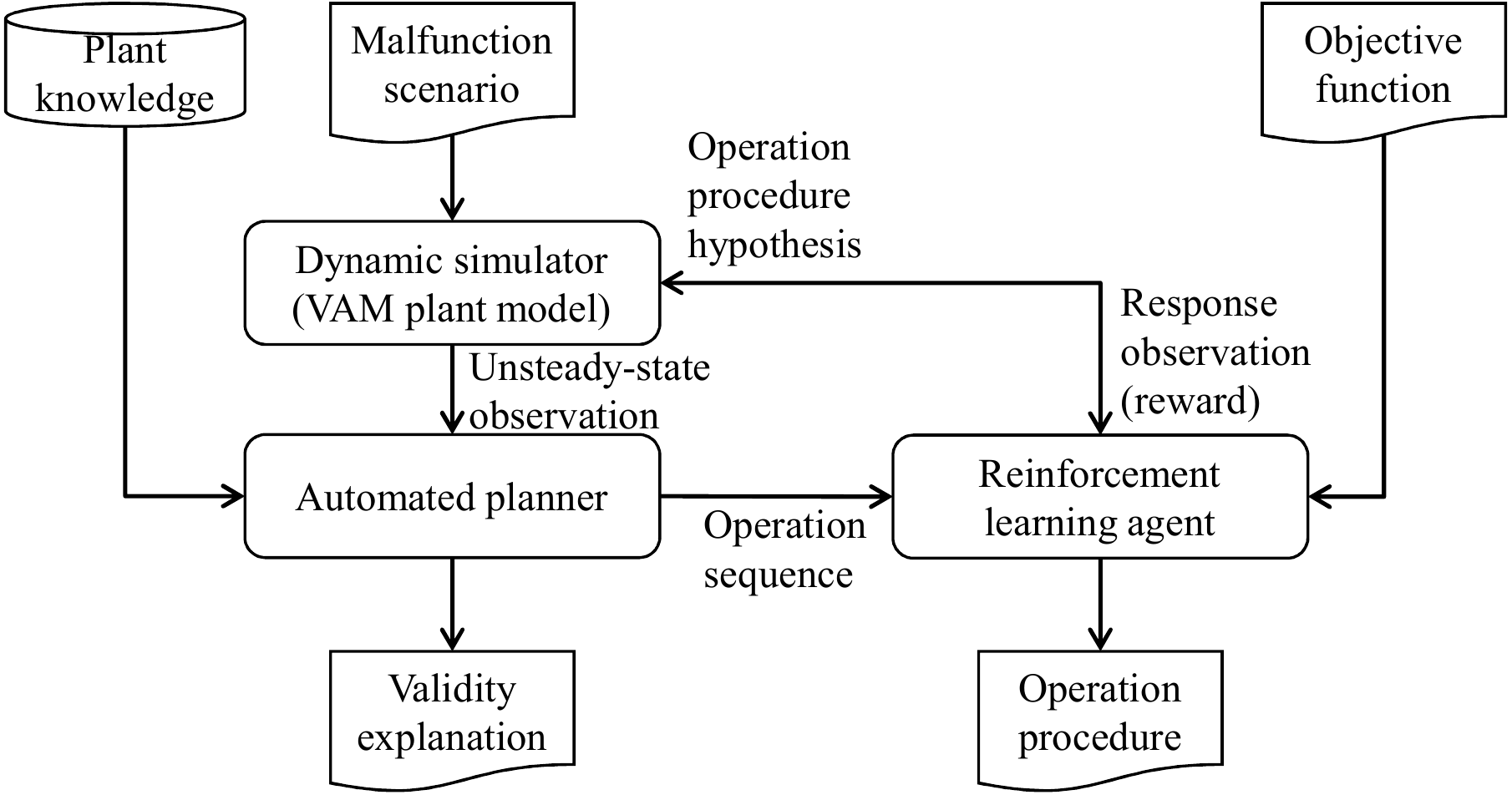}
\par\end{centering}
\caption{Block diagram of the proposed system\label{fig:Process-flow-of}}
\end{figure}

\section{Backgrounds}

\subsection{Classical planning}

Classical planning is a field of AI that studies how to explore action
sequences from one state to another based on a discrete state representation
and a discrete action (state transition rule) set. Classical planning
techniques can derive a sequence of elements to be manipulated using
qualitative behavioral models of a plant and has been applied to operation
procedure synthesis for chemical plants \cite{Aylett}.

\subsection{Reinforcement learning}

Reinforcement learning is a branch of machine learning that studies
how an \textit{agent} can aquire action sequences which maximize the
total reward given by an environment. Unlike classical planning, reinforcement
learning covers continuous control problems, and has been applied
to the problem of optimizing parameters of PID controllers \cite{Sedighizadeh}．

In order to deal with complex problems where no hand-engineered input
features are available, Deep Q-Networks (DQNs) \cite{Minh} combine
deep learning and reinfocement learning and achieved higher performance
than humans in the domain of video games. In reinforcement learning,
deep neural networks are often used as function approximators for
policy and value functions of the agent. However, the high representation
power of deep neural networks often results in unstable performance
due to its drastic changes in behavior while updating parameters.
Proximal Policy Optimization (PPO) \cite{PPO} is a relatively recent
deep reinforcement learning algorithm and has outperformed other state-of-the-art
algorithms on a virtual robot control problem on physical simulators
by introducing constraints for the amount of parameter changes on
each update to avoid the instability problem. 

\subsection{Dynamic simulation}

Dynamic simulation is used for reproducing behaviour and responses
of a dynamical system based on physical models. It has been applied
for the purpose of training operators \cite{Kano} and optimizing
operation procedures of industrial process plants \cite{Toshiba}.
The VAM plant model \cite{VAM,Omega} is one of those simulation enviroments.

\subsection{Procedure synthesis}

Automatic synthesis of operation procedures is a challenging theme
of plant control and has long been studied \cite{AOPS}. A recent
study \cite{Gofuku} uses a qualitative functional model based on
Multilevel Flow Modeling (MFM) for a plant and derives procedures
for leading a plant to a desired state by tracing back the influence
propagation rules between plant elements such as valves and pipes.
Since this type of method focuses on discrete qualitative models,
it cannot deal with continuous manipulation values of each operation
element and optimize them.

\section{Proposed system}

\begin{table}[t]
\caption{Examples of the plant knowledge\label{tab:knowledge}}

\centering{}%
\begin{tabular}{|c|}
\hline 
{\footnotesize{}IF fresh ethylene feed pressure increases} \\ {\footnotesize{}THEN
flow on FI101 (flowmeter) increases}\tabularnewline
\hline 
{\footnotesize{}IF pressure of V130 (vaporizer) increases }\\ {\footnotesize{}THEN
PC130 (PID) controls to close PCV101 (valve)}\tabularnewline
\hline 
{\footnotesize{}IF liquid level of V130 (vaporizer) increases} \\
{\footnotesize{}THEN LC130 (PID) controls to close LCV130 (valve) }\tabularnewline
\hline 
\end{tabular}
\end{table}
We develop an AI system that can automatically synthesize operation
procedures for chemical plants. It is designed to assist human operators
who are working to get the plant to recover from an undesirable (malfunction)
situation. Our system uses both qualitative and quantitative (dynamic
simulation) models of a plant and outputs operation procedures including
actual manipulation values that results from automatic exploration
of maximizing specified objective functions (rewards). The system
also outputs explanations about the validity of the procedures that
are human-understandable.

Figure \ref{fig:Process-flow-of} shows the block diagram of our system.
It consists of three modules, namely, the dynamic simulator of a plant,
the automated planner and the reinforcement learning agent. The planner
explores operation sequences and the agent explores the amount of
manipulation on each step of the sequence by interacting with the
simulator. The detailed process flow is as follows: 
\begin{enumerate}
\item The users of the system first prepare a knowledge base (qualitative
models) of the target plant and provide it to the automated planner.
Table \ref{tab:knowledge} describes an example of the knowledge.
Users also set the malfunction scenario to configure the undesirable
state on the dynamic simulator, and design the objective function
to be satisfied by the output procedure as the rewards of reinforcement
learning.
\item The dynamic simulator reproduces the undesired state according to
the scenario.
\item Given the observed undesirable state on the simulator, the automated
planner synthesizes operation sequences, i.e. which component to manipulate
in what order. The planner also outputs an explanation as to why the
sequence is valid. The explanation consists of a reasoning process,
utilized knowledge and the corresponding parts of the reference documents.
\item The reinforcement learning agent hypothesizes an operation procedure,
tries it on the simulator and receives sensor values as its response.
The agent considers a plan to be better if it yields greater rewards
calculated from observed sensor values. The agent repeats this exploration
many times until a certain termination condition, e.g. a threshold
of cumulative reward, is met.
\item The human plant operator decides whether to adopt the procedure output
by the system or not according to the explanation of its validity.
\end{enumerate}
This method utilizes knowledge not only for composing the explanation
but also for narrowing the space of possible procedures in order to
explore them in a practical computational time.

At the time of this writing, the reinforcement learning agent connected
to the dynamic simulator automatically explores operation procedures.
However, the module for automated planning is manually emulated.

\section{Preliminary experiments}

We conducted preliminary experiments to evaluate the efficiency of
the synthesized procedures for a recovery from a malfunction.

We used the VAM plant model \cite{VAM,Omega} as the dynamic simulator.
The materials used by a VAM plant are acetic acid, ethylene and oxygen.
They are processed through eight sections: (1) a raw material feed,
(2) a reactor, (3) a separator and a compressor, (4) an absorber,
(5) a CO2 remover and a purge line, (6) a buffer tank, (7) a distillation
column, and (8) a decanter. There are recycling systems of ethylene
and acetic acid, and the behavior of the plant is dynamic and complex.
The model has 36 malfunction scenarios, and they can reproduce unsteady
states caused by disturbances or process failures such as feed composition
change, heavy rain and failure of pumps.

We used the malfunction scenario, ``MAL03 Change C2H4 Feed Pressure''
(raw material ethylene (C2H4) feed pressure change, hereinafter called
MAL03) available on the plant model. MAL03 mainly affects the raw
material feed section on the plant. Figure \ref{fig:pnid} shows the
related components of the section. The PID controller, which monitors
the pressure of the vaporizer, controls the flow volume of the fresh
ethylene by adjusting the control valve. In this malfunction scenario,
the goal is to maintain the pressure of the vaporizer while the feed
pressure of the fresh ehylene changes. The occurrence of the malfanction
can be detected by observing the flowmeter attached on the fresh ethylene
feed pipe. By manual reasoning based on the plant knowledge base and
the qualitative observations, we specified one target element for
manipulation, which is the set point value (SV) of the PID controller.

In the following experiments, the reinforcement learning agent observes
real-valued seven dimensional vector of the sensor values around the
section and acts to decide the continuous scalar value of the SV at
each time step.

\subsection{Simple malfunction}

We compared the output procedure to standard PID control with respect
to the time required for recovery from malfunction.

The parameters of MAL03 are fixed through this experiment and the
change in the ethylene feed pressure (increase to 120\%) caused by
the malfunction occurs as a step function. The objective for the agent
to minimize is the difference between the value of the sensor on each
time and that of a normal state. We translated this into the reward
function as $\max(0,1-a|s_{t}-\sigma|)$ where $a$ is a constant
scale factor (we set it to $50$ in this experiment), $s_{t}$ is
the value of the sensor at time $t$ and $\sigma$ is the value of
the normal state. An episode (each procedure trial) consists of a
30-minutes simulation and actions executed on every one minute. The
agent receives a reward after each action. Thus, the cumulative reward
would be 30 if the plant kept steady during the whole episode, i.e.
if $s_{t}=\sigma$ for all $t$. We employed PPO as the reinforcement
learning agent.
\begin{figure}[t]
\begin{centering}
\includegraphics[width=7.5cm]{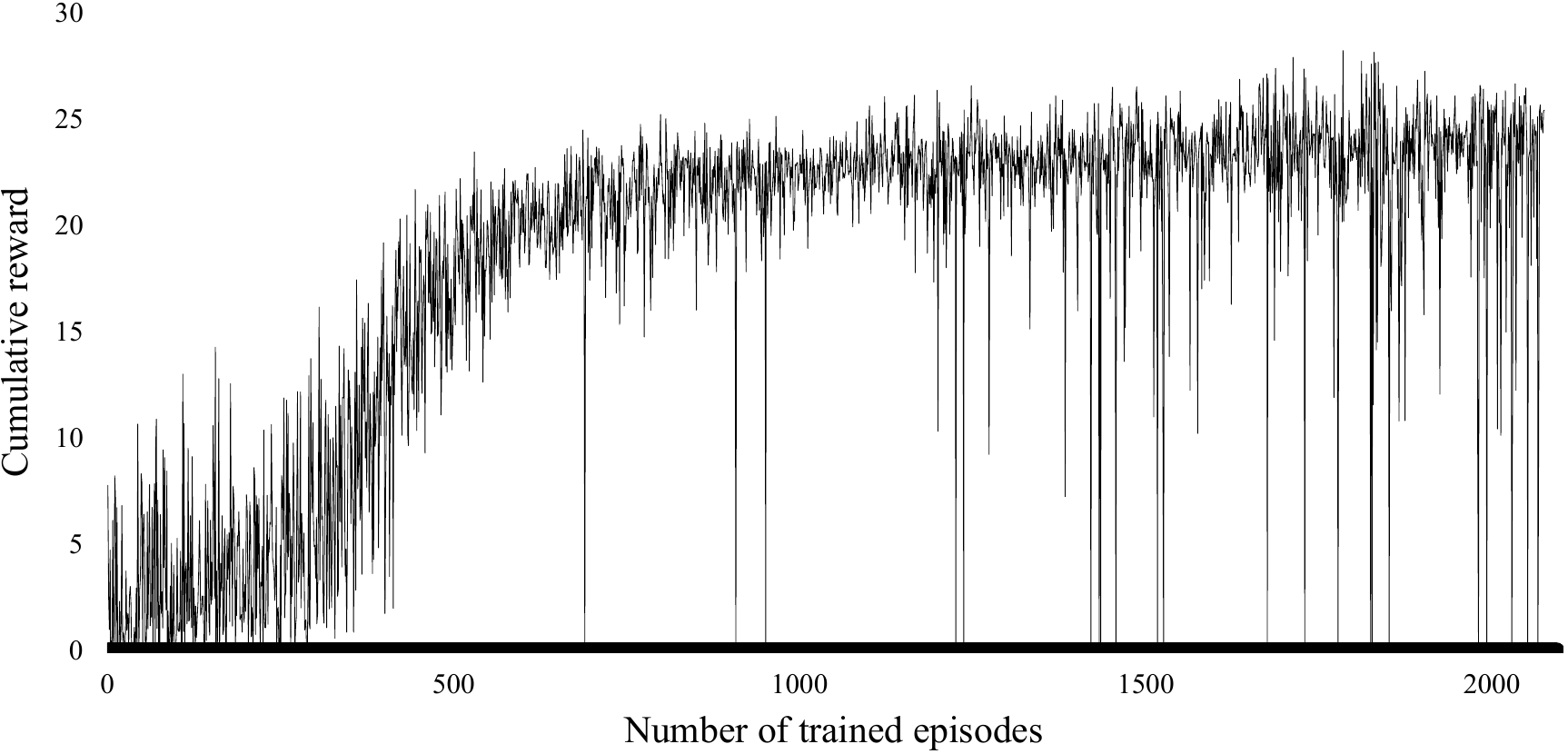}
\par\end{centering}
\caption{Training performance of PPO on the fixed malfunction with respect
to cumulative rewards\label{fig:mal03_simple}}
\end{figure}
\begin{figure}[t]
\begin{centering}
\includegraphics[width=7.5cm]{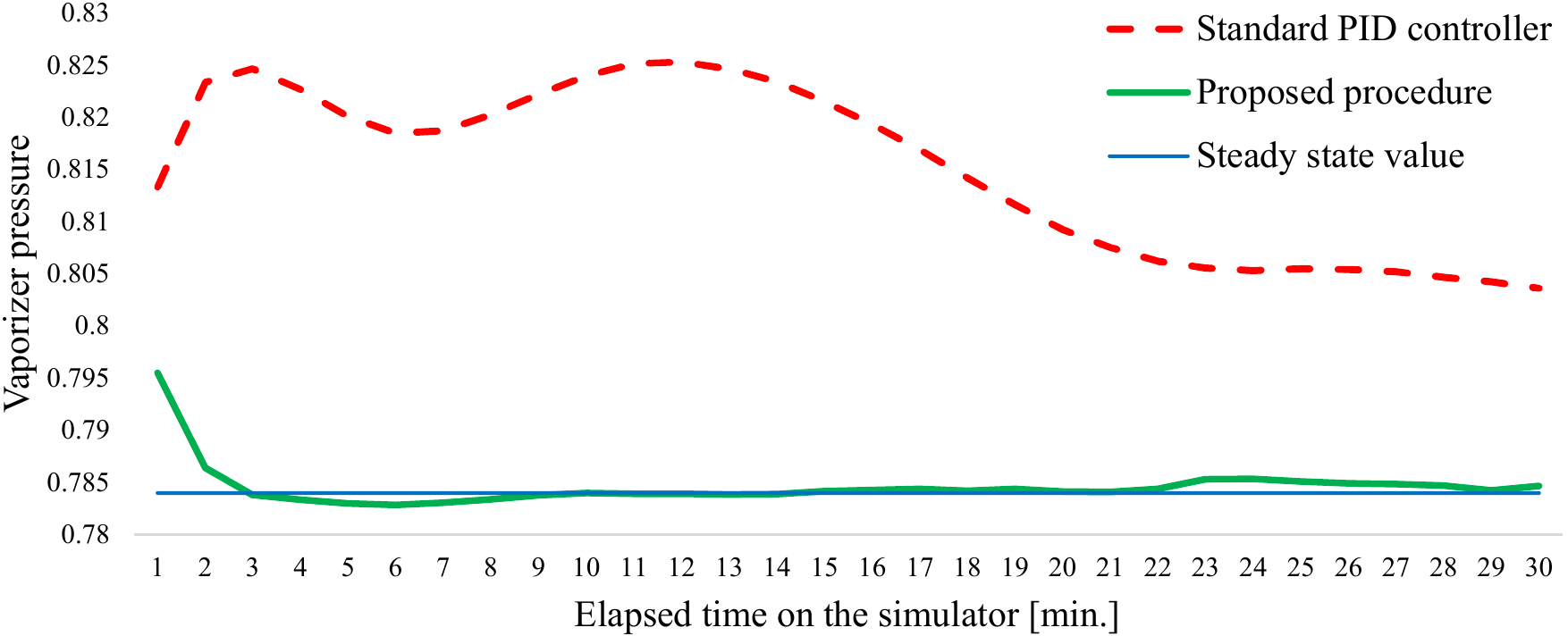}
\par\end{centering}
\caption{Simulated values of vaporizer pressures maintained by the proposed
procedure and the standard PID controller on the fixed malfunction
\label{fig:mal03_simple-pv}}
\end{figure}

Figure \ref{fig:mal03_simple} shows the cumulative rewards that the
agent received while the learning progresses. The thick black line
located at the bottom indicates the cumulative reward value of standard
PID control. This graph shows that the cumulative reward the agent
achieved after learning is much higher than in the case of standard
PID control.

Figure \ref{fig:mal03_simple-pv} shows the graphs of the target sensor
value (pressure of the vaporator) on the simulator. The straight horizontal
line (the value of 0.784) indicates the value of the normal state.
While the dashed line which indicates the response of the standard
PID controller does not achieve the steady state in the 30-minute
simulation, the proposed procedure quickly recovers from the unsteady
state in a few minutes. In other words, the agent successfully acquired
a procedure that enables a faster recovery from the malfunction. 

\subsection{Variable malfunction}

\begin{figure}[t]
\begin{centering}
\includegraphics[clip,width=7.5cm]{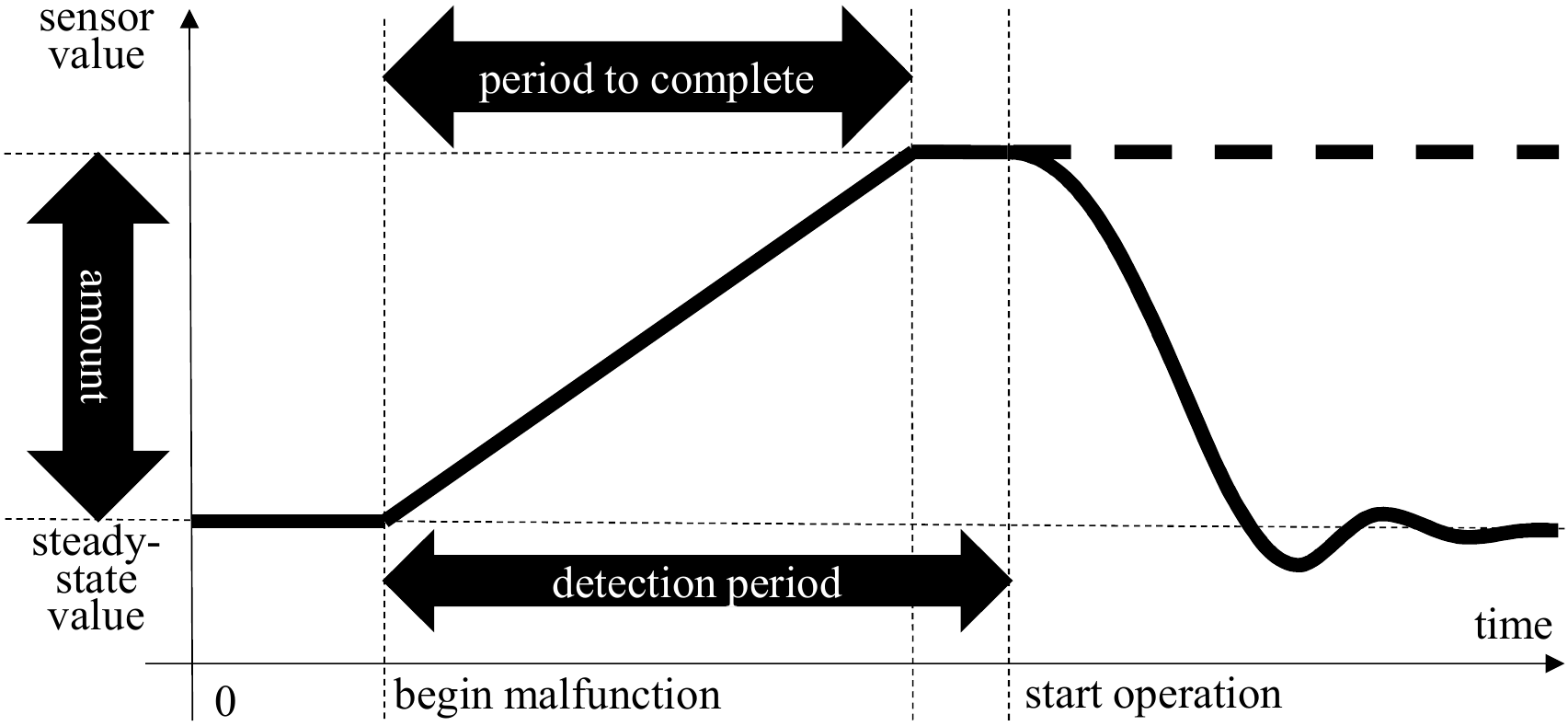}
\par\end{centering}
\caption{Set of parameters configured on the complicated MAL03 malfunction\label{fig:malconf}}
\end{figure}

We also evaluated the response ability of PPO in more complicated
settings. Figure \ref{fig:malconf} shows the set of parameters on
MAL03 which occurs as a ramp function. In this case, there are three
parameters, namely, the amount of difference from steady-state value,
the period to complete the malfunction and the period to start procedure
(depicted as thick arrows on the figure). 
To make MAL03 more complicated, we randomized the malfunction parameters
for each episode. The ranges of each parameter uniformly sampled from
are as follows:
\begin{itemize}
\item the amount of difference: from 90\% to 120\%,
\item the period to complete: 0 min. to 30 min.,
\item the period to start procedure: 0 min. to 60 min.
\end{itemize}
Other settings are almost the same as the simple case (simulate for
30 minutes). We also used PPO as the agent. 

\begin{figure}[t]
\begin{centering}
\includegraphics[width=7.5cm]{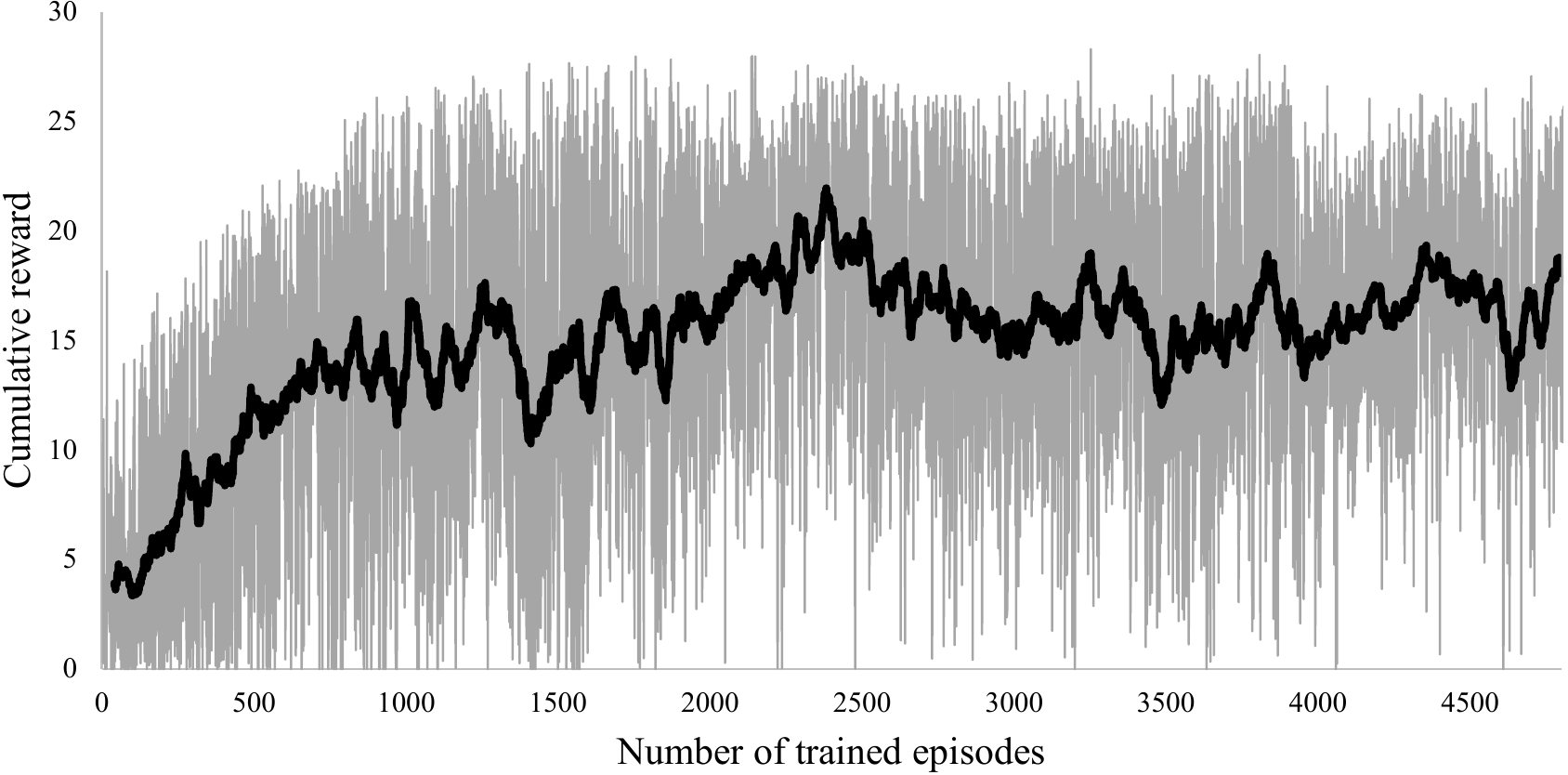}
\par\end{centering}
\caption{Training performance of PPO on the complicated malfunction with respect
to cumulative rewards\label{fig:mal03_complex}}
\end{figure}

Figure \ref{fig:mal03_complex} shows the cumulative reward during
the training of the agent. The bold line indicates the moving average
of 20 episodes. The reward value tends to increase along with the
number of episodes. This suggests that PPO allows the agent to learn
procedures for dynamic and complex ramp-wise disturbance which is
not well suited for MPC since it assumes step-wise changes. 

\section{Conclusion}

We have introduced an AI system for synthesizing plant operation procedures,
which we are developing for assisting human plant operators. The inputs
of the system are plant knowledge, a scenario for leading the plant
to an undesirable state, and an objective function. It outputs operation
procedures including detailed manipulation values and explanations
of the validity of the procedure that are understandable to human
operators. We have verified the efficiency of the output procedure
by preliminary experiments. We plan to evaluate the optimality of
the output procedure and the utility of explanation in collaboration
with plant operators as the next step of this work.

\end{document}